\documentclass[runningheads]{llncs}
\usepackage[T1]{fontenc}
\usepackage{graphicx}

\usepackage{latexsym}
\usepackage{amssymb}
\usepackage{amsmath}
\usepackage{booktabs}
\usepackage{enumitem}
\usepackage{graphicx}
\usepackage{color}
\usepackage{xurl}
\usepackage{hyperref}
\usepackage{orcidlink}

\usepackage{float}

\usepackage{xeCJK}
\usepackage{mdframed}
\usepackage{amsmath}
\usepackage{xcolor}
\usepackage{multirow}
\usepackage{changepage}

\newmdenv[
  backgroundcolor=white,       
  linecolor=gray!50,           
  linewidth=0.5pt,               
  frametitlebackgroundcolor=gray!50, 
  frametitlefont=\color{black}\bfseries, 
  roundcorner=10pt,                  
  innertopmargin=10pt,               
  innerbottommargin=10pt,            
  innerrightmargin=10pt,             
  innerleftmargin=10pt,              
  skipabove=10pt,                    
  skipbelow=10pt                     
]{promptbox}

\begin{document}
%
\title{A Comparative Study of Demonstration Selection for Practical Large Language Models-based Next POI Prediction}
\titlerunning{Demonstration Selection for Next POI Prediction}
%
\author{Ryo Nishida\orcidlink{0000-0001-8444-0852} \and
Masayuki Kawarada\orcidlink{0009-0009-2467-9429} \and 
Tatsuya Ishigaki\orcidlink{0000-0003-3278-2345} \and
Hiroya Takamura\orcidlink{0000-0002-3244-8294} \and 
Masaki Onishi\orcidlink{0000-0002-4580-4868}
}

\authorrunning{R. Nishida et al.}
%
\institute{National Institute of Advanced Industrial Science and Technology, Japan 
\email{\{ryo.nishida, kawarada.masayuki, ishigaki.tatsuya, \\ takamura.hiroya, onishi-masaki\}@aist.go.jp}}
\maketitle              

\begin{abstract}
This paper investigates demonstration selection strategies for predicting a user's next point-of-interest (POI) using large language models (LLMs), aiming to accurately forecast a user's subsequent location based on historical check-in data.
While in-context learning (ICL) with LLMs has recently gained attention as a promising alternative to traditional supervised approaches, the effectiveness of ICL significantly depends on the selected demonstration.
Although previous studies have examined methods such as random selection, embedding-based selection, and task-specific selection, there remains a lack of comprehensive comparative analysis among these strategies.
To bridge this gap and clarify the best practices for real-world applications, we comprehensively evaluate existing demonstration selection methods alongside simpler heuristic approaches such as geographical proximity, temporal ordering, and sequential patterns. 
Extensive experiments conducted on three real-world datasets indicate that these heuristic methods consistently outperform more complex and computationally demanding embedding-based methods, both in terms of computational cost and prediction accuracy. 
Notably, in certain scenarios, LLMs using demonstrations selected by these simpler heuristic methods even outperform existing fine-tuned models, without requiring further training.
Our source code is available at: \url{https://github.com/ryonsd/DS-LLM4POI}.

\end{abstract}

\section{Introduction}

Next point-of-interest (POI) prediction is essential for location-based services such as Google Maps~\footnote{\url{https://www.google.com/maps}} and Foursquare~\footnote{\url{https://foursquare.com}}, enabling systems to anticipate users' future locations based on historical check-in data. Unlike traditional recommendation tasks, next POI prediction uniquely incorporates rich spatiotemporal signals—such as location-specific routines and time-of-day preferences—reflecting complex human mobility patterns.

Conventional approaches typically employ supervised models, including graph neural networks and sequence-to-sequence architectures~\cite{Luo_STAN_WWW2021,Yang_GETNext_SIGIR2022,wang_adaptive_graph_SIGIR2023,Yan_STHGCN_SIGIR2023}, which, although effective, demand costly retraining and extensive labeled data. This limits their flexibility in real-world scenarios characterized by constantly evolving user behaviors and the frequent addition of new POIs~\cite{Cai04032018}.
In response to these limitations, in-context learning (ICL) with large language models (LLMs) has emerged as an attractive alternative~\cite{wang2023zeroshotnextitemrecommendationusing,geng_P5_KDD2023,wu2024surveylargelanguagemodels}. ICL enables LLMs to make predictions without fine-tuning by conditioning on a limited set of examples (demonstrations) provided in the prompt. Consequently, demonstration selection critically influences performance, as ICL relies on effectively identifying the most informative examples.
However, demonstration selection strategies for next POI prediction remain underexplored. Existing works predominantly apply methods borrowed from natural language processing (NLP)~\cite{peng-etal-2024-revisiting,kawarada-etal-2024-demonstration,zhu-etal-2024-towards-robust}, such as random selection and task-specific selection~\cite{wang_LLMMob_2024} or computationally intensive embedding-based selection~\cite{Li_LLM4POI_SIGIR2024}. These methods often fail to adequately capture the unique spatiotemporal dynamics intrinsic to human mobility data, and embedding-based methods, in particular, suffer from scalability issues due to high computational complexity.

To bridge existing research gaps and identify best practices for real-world applications, this study comprehensively investigates demonstration selection specifically tailored for next POI prediction. 
We propose three intuitive, computationally efficient heuristic-based strategies that directly leverage spatiotemporal similarities: a spatial approach based on geographic proximity, a set-based approach treating check-ins as unordered sets, and a sequence-based approach preserving temporal order.
Additionally, we rigorously examine a foundational yet understudied aspect: the optimal composition of the demonstration pool, comparing the use of demonstrations from the target user's own history versus those drawn from all users.


Our experiments compared these selection strategies across three real-world datasets: Foursquare-New York, Foursquare-Tokyo, and Gowalla-California, which feature diverse check-in records from different cities and platforms.
The results show that heuristic-based strategies based on spatial, set, and sequence similarity consistently outperform both embedding-based and random selection. Moreover, these heuristic-based methods have a lower computational cost during the selection phase compared to embedding-based selection. The performance advantage is particularly significant in scenarios with fewer demonstrations, where selecting the most informative examples is critical due to context length limitations. Notably, the top-performing strategy achieves results competitive with those of fine-tuned models, but without requiring any model training.



The remainder of this paper is organized as follows.
Section 2 reviews related works of ICL in general recommendation tasks and demonstration selection.
Section 3 describes the task definition, the prompt of LLMs, and simpler heuristic demonstration selection strategies.
Section 4 presents the experimental setup and baselines.
Section 5 discusses the results and analysis.
Finally, Section 6 concludes the paper and outlines future work.

\section{Related Work}

We review related work in two main areas relevant to our study. First, we consider the use of ICL in recommender systems, since next POI prediction can be viewed as a personalized recommendation task over spatiotemporal data. Second, we examine demonstration selection strategies in ICL-based recommendation tasks.

\subsection{ICL in Recommendation}
ICL has recently gained momentum in studies of recommender systems.
This paradigm has been applied in recommender systems for various domains such as movie, music, and news recommendation. For example, prior research has demonstrated its effectiveness in zero-shot next-item recommendation~\cite{wang2023zeroshotnextitemrecommendationusing}, in surveys of LLMs for recommendation~\cite{wu2024surveylargelanguagemodels}, and in domain-specific applications~\cite{Dai2023}. Recent studies further show that prompting LLMs with a user's historical consumption data and relevant item information can lead to strong recommendation performance, sometimes even rivaling fine-tuned models.

Traditional methods for next POI prediction include approaches such as Markov Chains~\cite{chen_where_ijcai2013}, recurrent neural networks~\cite{Sun2020-ap}, transformer-based architectures~\cite{Yang_GETNext_SIGIR2022}, and graph neural networks that capture sequential and spatial dependencies from check-in histories~\cite{Yan_STHGCN_SIGIR2023}.

While effective, these methods often incur high training costs, as they require task-specific model training on large datasets. To overcome these limitations, recent studies on next POI prediction have started to explore the use of LLMs. For instance, researchers have fine-tuned LlaMA2~\cite{touvron2023llama2openfoundation} for POI prediction and demonstrated that LLM-based methods can outperform traditional architectures~\cite{Li_LLM4POI_SIGIR2024}. Moreover, ICL has also been applied directly to next POI prediction, bypassing the need for fine-tuning~\cite{shanshan_whereto_cai2024,wang_LLMMob_2024}.

\begin{figure}[t]
\centering
\includegraphics[width=\linewidth]{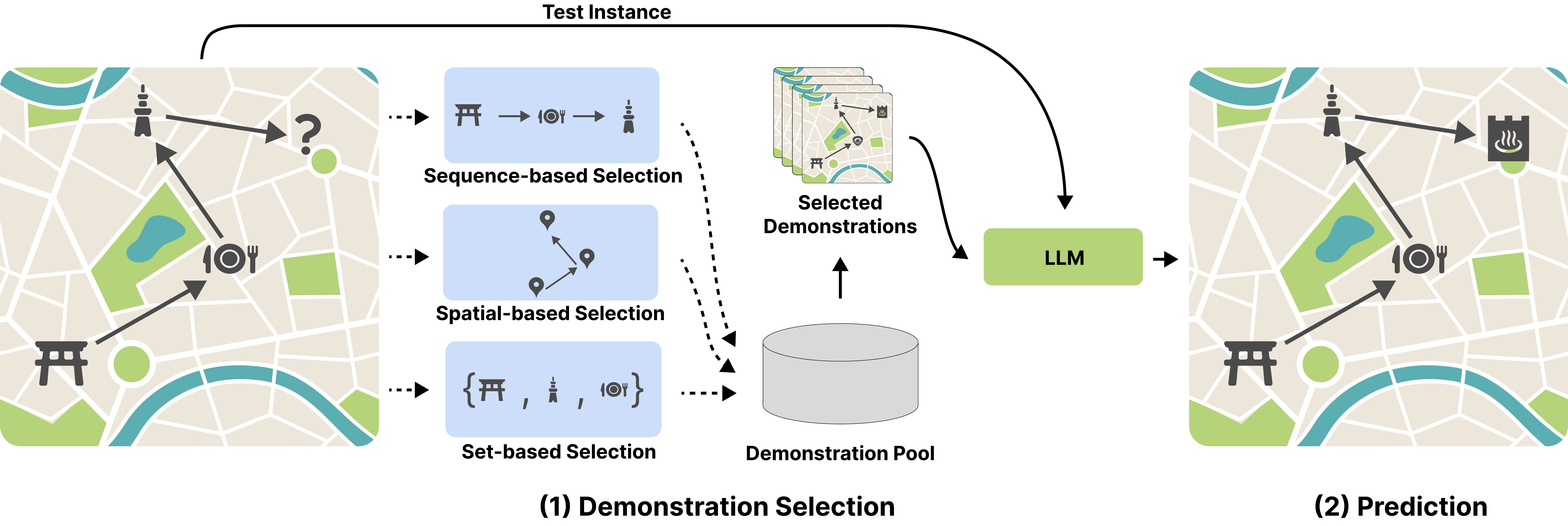}
\vspace{-0.5cm}
\caption{An overview of the next POI prediction task and an overview of heuristic-based demonstration selection methods. This task is to predict the next POI the user is likely to visit, given a sequence of users' check-in records.
The heuristic-based methods (1) retrieve similar instances from a demonstration pool (i.e., past check-in records) and (2) predict the next POI including them in a prompt as demonstrations.}
\label{fig:method_jp}
\end{figure}

\subsection{Demonstration Selection}

Demonstration selection is a central research topic in studies of ICL.
Many NLP studies have proposed strategies for selecting demonstrations. Common approaches include random sampling or embedding similarity-based selection~\cite{peng-etal-2024-revisiting,kawarada-etal-2024-demonstration,zhu-etal-2024-towards-robust}.
Domain-specific demonstration selection strategies have been also proposed for recommendation tasks.
For example, it is shown that in domains such as movie, music, and game recommendation, using retrieved demonstrations based on item overlap or semantic similarity significantly outperforms random selection~\cite{wang-lim-2024-whole}.
These findings highlight that selecting informative demonstrations is crucial also in recommendation tasks.
Although next POI prediction can be a recommendation task, demonstration selection remains underexplored for this task. Existing ICL-based methods rely on simple heuristics—for instance, selecting check-in records from a fixed number of recent days~\cite{wang_LLMMob_2024,shanshan_whereto_cai2024}. Such heuristics are straightforward but may fail to capture important spatial and behavioral patterns in user mobility.

In contrast to many NLP tasks—where demonstrations consist of tokens of natural language—POI data involves structured inputs with geospatial coordinates, timestamps, and category IDs. This makes applying standard embedding similarity-based retrieval techniques from NLP non-trivial. Prior studies in NLP have investigated demonstration selection in tasks like document classification, question answering, and language generation from structured data~\cite{kawarada-etal-2024-demonstration,kawarada-etal-2024-prompting-numerical}, but these methods do not directly account for spatiotemporal context. In this study, we tackle this gap by evaluating several demonstration selection strategies tailored to next POI prediction.

\section{Methods}

Our goal is to predict the next POI a user will visit during specific periods (e.g., trips or holidays). 
This section defines the task, presents the prompt, and shows simpler heuristic demonstration selection strategies.

\subsection{Task Definition}

We aim to predict a user’s next POI based on their check-in history. Each check-in is represented as \((u, p, c, t, \mathbf{g})\), where \(u\) represents a user, \(p\) represents a POI, \(c\) denotes the POI category, \(t\) is the timestamp, and \(\mathbf{g}\) indicates the 2D geographical location (latitude, longitude).
Let $D$ be the set of check-in records of all users in a location-based service.

The check-in history of a user \( u \) is denoted as:
\[
\mathfrak{D}^u = \{ (p_1, c_1, t_1, \mathbf{g}_1), \dots, (p_J, c_J, t_J, \mathbf{g}_J) \}.
\]

In this task, we focus on predicting the next POI a user will visit during specific periods as mentioned earlier.
To account for temporal structures, the check-in history \( \mathfrak{D}^u \) is divided into sequences based on a time interval \( \Delta t \).
Each sequence represents a user's visit pattern over a defined period:
\[
\mathfrak{D}^u = \{ \mathbf{T}^u_1, \mathbf{T}^u_2, \dots, \mathbf{T}^u_S \},
\]
where \( \mathbf{T}^u_s \) is the \( s \)-th sequence of check-ins, and \( S \) is the total number of sequences for user \( u \).
Each sequence consists of consecutive check-in records: $
\mathbf{T}^u_s = \{ (p_{i_1}, c_{i_1}, t_{i_1}, \mathbf{g}_{i_1}), \dots, (p_{i_L}, c_{i_L}, t_{i_L}, \mathbf{g}_{i_L}) \}.$
Here, \( \{i_1, \dots, i_L\} \) represents the check-ins belonging to the sequence \( s \).
We define \( \mathfrak{D} \) as the merged set of all user check-in histories: $\mathfrak{D} = \cup_{u \in U} \mathfrak{D}^u.$

The next POI prediction is defined as the task to predict the last $p_{i_{L}}$ in the user $u$'s history \( \mathbf{T}^{u}_s\), given a timestamp $t_{iL}$.

We use two demonstration pools: $\mathfrak{D}^u$ and $\mathfrak{D}$.
$\mathfrak{D}^u$ is the set of past visits of the target user, and reflects the personal movement patterns of the target user. $\mathfrak{D}$ is the set of past visits of all users, and captures general mobility trends across different individuals.
By selecting relevant demonstrations from these pools, we aim to provide the model with useful contextual information, improving its ability to predict the next POI more accurately.

\subsection{Prompt}
We focus on ICL for the next POI prediction task. In this approach, a prompt is constructed to leverage the knowledge in LLMs for predicting the next POI a user is likely to visit.
Figure~\ref{fig:prompt} illustrates the prompt structure, which includes the task instruction, demonstrations, and the test instance.
The task instruction follows previous work~\cite{wang_LLMMob_2024} and defines the goal: ``{\it Your task is to predict a user’s next location based on his/her activity pattern.}’’
In addition, we explicitly specify the aspects that should be considered for prediction, e.g., ``{\it considering the following aspects: 1. the activity pattern of this user that you learned from examples, e.g., repeated visits to
a certain place during a certain time}... (omitted)''.

Demonstrations consist of user trajectories and corresponding next POIs, as shown in Figure~\ref{fig:prompt}. Each check-in is represented in the following format: (check-in time, day of the week, POI ID, POI category).
The check-in time is expressed in 12-hour format.
For a given check-in sequence, all check-ins except the last one are used as the context, and the final check-in is treated as the target. Together, they form a single demonstration example. Note that although POI names could provide additional semantic cues, we do not incorporate them in this study, as they are not available in most public POI datasets and to ensure fair comparison with prior work.

The test instance follows the same format, with the current trajectory (<context\_current>) and the target location (<target\_current>), from which the model predicts the next POI.

\begin{figure}[t]
\centering
\begin{promptbox}[frametitle=Prompt Template]
\ttfamily
\fontsize{7pt}{0cm}\selectfont
Your task is to predict a user's next location based on his/her activity pattern. You will be provided with some examples of the user's historical stay sequences.
One sequence consists of <context> and <target>. <context> provides contextual information about where and when this user has been to before the last stay. <target> is the last stay in the sequence. Stays in <context> are in chronological order.
Each stay takes on such form as (start\_time, day\_of\_week, place\_id, place\_category). 

The detailed explanation of each element is as follows:

\begin{adjustwidth}{1em}{0em} start\_time: the start time of the stay in 12h clock format.
\end{adjustwidth}

\begin{adjustwidth}{1em}{0em} day\_of\_week: indicating the day of the week.
\end{adjustwidth}

\begin{adjustwidth}{1em}{0em} place\_id: an integer representing the unique place ID, which indicates where the stay is.
\end{adjustwidth}

\begin{adjustwidth}{1em}{0em} place\_category: a string representing the category of the place (e.g., Train Station, Park, etc.).
\end{adjustwidth}

<target\_current> is the prediction target with unknown place ID denoted as <next\_place\_id> and unknown place category name denoted as <next\_place\_category>, while temporal information is provided.     

Please infer what the \textless next\_place\_id\textgreater\ is (i.e., the most likely place ID), considering the following aspects:
    
\begin{adjustwidth}{1em}{0em} 1. the activity pattern of this user that you learned from examples, e.g., repeated visits to a certain place during a certain time;
\end{adjustwidth}

\begin{adjustwidth}{1em}{0em} 2. the context stays in \textless context\_current\textgreater, which provide more recent activities of this user; 
\end{adjustwidth}

\begin{adjustwidth}{1em}{0em} 3. the temporal information (i.e., start\_time and weekday) of target stay, which is important because people's activity varies during different time (e.g., nighttime versus daytime) and on different days (e.g., weekday versus weekend).
\end{adjustwidth}

Please organize your answer in a JSON object containing following keys: "place\_id" and "place\_category". Do not include reasons in your output.

\textcolor{blue}{
The examples are as follows:
\begin{adjustwidth}{1em}{0em} <context>: (01:22 PM, Wednesday, 2436, Train Station), (09:08 AM, Thursday, 3544, Gym / Fitness Center)
\end{adjustwidth}
\begin{adjustwidth}{1em}{0em}
    <target>: (00:13 PM, Thursday, 3824, Department Store)
\end{adjustwidth}
$\cdots$
}

\textcolor{orange}{
The current data are as follows:
\begin{adjustwidth}{1em}{0em} <context\_current>: (00:39 PM, Wednesday, 480, Department Store), (02:52 PM, Wednesday, 1218, Coffee Shop)
\end{adjustwidth}
\begin{adjustwidth}{1em}{0em}  <target\_current>: (10:13 AM, Thursday, <next\_place\_id>, <next\_place\_category>)
\end{adjustwidth}
}

\end{promptbox}
\vspace{-0.4cm}
\caption{Prompt template. Task instruction (black), demonstrations (blue), and input as a test instance (orange).}
\label{fig:prompt}
\end{figure}

\subsection{Demonstration Selection}

This section describes how we select demonstrations (the blue part in the prompt in~Figure~\ref{fig:prompt}).  
Although previous studies have examined methods such as random selection, embedding-based selection, and selecting check-in sequences from
a fixed number of recent days~\cite{shanshan_whereto_cai2024,wang_LLMMob_2024,Li_LLM4POI_SIGIR2024}, there remains a lack of comprehensive comparative analysis among these strategies. Therefore, we comprehensively evaluate existing demonstration selection methods alongside the new simpler heuristic approaches.
The key motivations behind the new approaches are: 1) users who have followed similar routes in the past are also likely to take similar routes in the future; additionally, 2) individual users tend to exhibit consistent movement patterns over time, meaning their future visits are often influenced by their own past behavior.
By selecting demonstrations that closely resemble the test instance, we can provide the LLM with more relevant contextual knowledge, thereby improving its ability to accurately predict the next POI.

\subsubsection{Simpler Heuristic Demonstration Selection Strategies}

We evaluate three demonstration selection strategies:
1) spatial-based selection, which selects demonstrations based on geographical proximity,  
2) set-based selection, which treats check-in records as an unordered set and selects those with overlapping POIs, and   
3) sequence-based selection, which considers the temporal order of check-ins and selects trajectories with similar visit sequences.

In few-shot prompting, we select \(k\) demonstrations $
\mathfrak{D}_{\mathbf{demo}} = 
\{
\mathbf{T}_{1}, 
\cdots,
\mathbf{T}_{k} 
\}
$
using one of the selection methods. These demonstrations are drawn either from the target user's history, \(\mathfrak{D}^{\mathbf{u}}\), or from all users' histories, \(\mathfrak{D}\), depending on the strategy.
Given an input instance \(\mathbf{T}^{\mathbf{u}}_s\), the selected demonstrations are included in the prompt.
The selection strategies are detailed below:

\begin{itemize}

 \item \textbf{Spatial-based Selection~(DTW):} We represent a user's check-in sequence as a time series of two-dimensional coordinates.
To measure spatial similarity between trajectories, we use Dynamic Time Warping (DTW)~\cite{conf/vldb/Keogh02}.
For each test instance, we compute the DTW distance between its check-in sequence and those in a demonstration pool.
The top $k$ sequences with the smallest DTW values are selected as demonstrations.

\item \textbf{Set-based Selection~(Jaccard):} We treat check-in records as sets of POI IDs and measure similarity using the Jaccard coefficient, which quantifies the proportion of shared POIs between two trajectories.  
The top $k$ sequences with the highest Jaccard similarity are selected as demonstrations.

\item \textbf{Sequence-based Selection~(LCS):}
We represent check-in records as temporally ordered sequences of POI IDs and compute similarity using the longest common subsequence (LCS).  
This method selects the top $k$ sequences with the longest matching subsequence of POIs, favoring trajectories where the same POIs appear in the same order.

\end{itemize}

\subsubsection{User-based Filtering Strategies}

We also introduce a user-based filtering strategy, where demonstrations are selected exclusively from the target user’s own past check-in history, $\mathfrak{D}^u$.
This approach ensures that the selected demonstrations closely reflect the user's personal movement patterns.
(\textbf{DTW+User}, \textbf{Jaccard+User}, and \textbf{LCS+User}). 
Only when we use the user-based filtering, we can use another selection method:
\begin{itemize}
 \item \textbf{Temporal-based Selection (Time+User):}  this retrieves demonstrations from $\mathfrak{D}^u$ in reverse chronological order, selecting the most recent check-in sequences first. This approach, previously used in LLM-Mob~\cite{wang_LLMMob_2024}, assumes that a user's most recent mobility patterns are the most relevant for predicting their next visit.  
\end{itemize}

By incorporating user-based filtering, we aim to evaluate how well LLMs leverage personal mobility patterns in next POI prediction, either in isolation or in combination with similarity-based selection strategies.

\section{Experiments}
We describe the datasets and the compared models.
We validate the effectiveness of several methods for selecting demonstrations from previously accumulated history in the next POI prediction using few-shot prompting with LLMs.

\subsection{Datasets}  
We use the Foursquare-New York, Foursquare-Tokyo~\cite{yang_zhan_modeling_ieee2015}, and Gowalla-California~\cite{cho2011friendship} datasets for the experiments. These datasets contain check-in data from the Foursquare location-based service in the New York City and Tokyo areas, and Gowalla in the California and Nevada areas. 
We apply the same preprocessing steps as in \cite{Li_LLM4POI_SIGIR2024} to ensure fair comparisons with existing models.
In particular, we segment a user's check-in history into separate sequences whenever there is a gap of 24 hours or more between consecutive check-ins, treating them as distinct movement patterns.  
Additionally, all users included in the test data have at least one historical check-in sequence in the training data.  
The statistics information of the each datasets are presented in Table~\ref{tab:dataset_stats}. 
Note that since a larger number of POIs makes the prediction task more challenging, the difficulty of next POI prediction increases in the order of NYC, TKY and CA.

\begin{table}[t]
  \caption{Statistics of datasets used in our experiments. NYC, TKY, and CA represent Foursquare-New York, Foursquare-Tokyo, and Gowalla-California, respectively.}
  \centering
  \begin{tabular}{lccc}
    \toprule
    & NYC & TKY & CA \\
    \midrule
    Number of Users & 1,048 & 2,282 & 3,957 \\
    Number of POIs & 4,981 & 7,833 & 9,690 \\
    Number of Test Instances & 1,447 & 7,079 & 2,864 \\
    \midrule
    Average Historical Data per User & 11.9 & 23.8 & 14.3 \\
    \bottomrule
  \end{tabular}
  \label{tab:dataset_stats}
\end{table}

\subsection{Compared LLMs and Baselines}\label{setting}  

We comprehensively evaluate demonstration selection methods using two types of LLMs:  
Qwen-2.5-7B-Instruct~\cite{qwen2.5}, which is an open-source LLM, and OpenAI's GPT-4o.
To benchmark our method, we compare it against the following baseline methods.

\subsubsection{Baseline Demonstration Selection Methods}

We use commonly adopted baselines in ICL for recommendation tasks, namely random selection and embedding-based selection. 
\textbf{Random}: is method that randomly selects $k$ demonstrations from the demonstration pools. We perform five trials for each setting and report the average results. 
\textbf{Embedding Similarity (EmbSim)}: selects demonstrations by computing cosine similarity between text-embedded check-ins. Each check-in is converted into a textual description and embedded using LlaMa2-7B~\cite{touvron2023llama2openfoundation}. 
The top $k$ most similar instances are selected, following the approach of Li et al.~\cite{Li_LLM4POI_SIGIR2024}.

\subsubsection{Existing Fine-Tuning and In-Context Learning Models}

We compare our ICL-based methods with fine-tuning and existing ICL models.
We use the following four models. Notably, all these models are trained on historical data from all users. The first two are deep learning models that were trained specifically for the task and are not based on LLMs.
\textbf{GETNext}~\cite{Yang_GETNext_SIGIR2022}: incorporates a global trajectory graph capturing POI transition patterns, along with user preferences and spatio-temporal contexts.
\textbf{STHGCN}~\cite{Yan_STHGCN_SIGIR2023}: models both intra-user and inter-user trajectory-level relations using a hypergraph. It integrates a hypergraph transformer to combine spatio-temporal information with higher-order collaborative signals.
The third model is based on an LLM and has been fine-tuned for the task.
\textbf{LLM4POI}~\cite{Li_LLM4POI_SIGIR2024}: converts check-in sequences into natural language prompts and uses a fine-tuned LlaMa2 to predict the next POI. Similar check-in sequences are included in the prompt based on embedding similarity, following a similar approach to \textbf{EmbSim}.
The fourth model is used as a baseline for comparison with an ICL-based approach that does not involve fine-tuning.
\textbf{LLM-Mob}~\cite{wang_LLMMob_2024}: The prompts are structured similarly to ours shown in Figure~\ref{fig:prompt}, however, LLM-Mob simply lists check-in records as tuples (e.g., ``(01:22 PM, Wednesday, 2236, Train Station)’’), and they are not divided history into segments and not represents as ``<context>'' and ``<target>''.
LLM-Mob selects demonstrations by \textbf{Time+User}.
 Comparing our methods with LLM-Mob allows us to analyze the effects of different representations of demonstrations in prompts.

\subsection{Experimental Settings}
To analyze the effect of the number of demonstrations on performance, we conduct experiments with different values of \( k \): 5, 15, and 30.  
If a user's past check-in sequence contains fewer than \( k \) records, only the available check-in sequences are used as demonstrations. This ensures the model does not receive more examples than exist.
We evaluate the prediction performance using a common metric, accuracy@1 (ACC@1), as commonly done in related work~\cite{Li_LLM4POI_SIGIR2024,wang_LLMMob_2024,shanshan_whereto_cai2024}.
This metric measures the proportion of test instances in which the predicted POI ID matches the actual next POI visited by the user. 

In addition to prediction accuracy, we also evaluate each demonstration selection method in terms of computational cost, focusing on the demonstration selection time required per test instance. Embedding-based selection methods require GPU acceleration to calculate embedding features and were run on a NVIDIA GeForce GTX TITAN X, while all other methods were executed on CPU only (Intel Core i9-9900K @ 3.60GHz).

\begin{table}[H]
\centering
\caption{ACC@1 scores for each demonstration selection method using historical data from all users (All) and using historical data from the target user only (User) using Qwen-2.5-7B-Instruct and GPT-4o. Random in the User column is equivalent to Random+User, and this correspondence holds for the other selection methods. The highest ACC@1 score is highlighted in bold for each of the four settings defined by the combination of model (Qwen-2.5-7B or GPT-4o) and historical data pool (All or User). Additionally, for each setting, we perform McNemar’s test to compare the performance of each method against Random or Random+User, and denote $^{\dagger}$ and $^{\ddagger}$ when $p < 0.1$ and $p < 0.05$, respectively.}
\centering
\small
\resizebox{\linewidth}{!}{
  \begin{tabular}{@{}c c c c c c c c c c c c@{}}
    \toprule
      &  &  
      & \multicolumn{3}{c}{\textbf{NYC}} 
      & \multicolumn{3}{c}{\textbf{TKY}} 
      & \multicolumn{3}{c}{\textbf{CA}} \\
    \cmidrule(lr){4-6} \cmidrule(lr){7-9} \cmidrule(lr){10-12}
      &  &  
      & 5 & 15 & 30  
      & 5 & 15 & 30  
      & 5 & 15 & 30  \\
    \midrule
    \multirow{12}{*}{\rotatebox[origin=c]{90}{Qwen-2.5-7B}}
      & \multirow{6}{*}{All}  
        & Random   
          & 0.1237 & 0.1313 & 0.1431  
          & 0.0743 & 0.0808 & 0.0930  
          & 0.0544 & 0.0583 & 0.0615  \\
      &       & EmbSim   
          & 0.1320 & 0.1361 & 0.1348  
          & 0.0824 & 0.0903 & 0.0898  
          & 0.1082 & 0.1094 & 0.1023       \\
      \cmidrule{3-12}
      &       & DTW      
          & 0.1500$^\dagger$ & 0.1548$^\dagger$ & 0.1513$^\dagger$  
          & 0.1305$^\ddagger$ & 0.1302$^\ddagger$ & 0.1302$^\ddagger$  
          & 0.0985      & 0.1192$^\dagger$      & 0.1209$^\ddagger$      \\
      &       & Jaccard  
          & \textbf{0.1790}$^\ddagger$ & 0.1679$^\ddagger$ & 0.1755$^\ddagger$  
          & 0.1382$^\ddagger$ & \textbf{0.1420}$^\ddagger$ & \textbf{0.1400}$^\ddagger$  
          & 0.1041      &  0.1235$^\ddagger$     &   0.1230$^\ddagger$     \\
      &       & LCS      
          & 0.1776$^\ddagger$ & \textbf{0.1804}$^\ddagger$ & \textbf{0.1783}$^\ddagger$  
          & 0.1291$^\ddagger$ & 0.1369$^\ddagger$ & 0.1321$^\ddagger$  
          & \textbf{0.1121}$^\ddagger$          &  \textbf{0.1279}$^\ddagger$     &  \textbf{0.1392}$^\ddagger$      \\
      \cmidrule{2-12}
      & \multirow{7}{*}{User}  
        & Random   
          & 0.1838 & 0.1983 & 0.1997  
          & 0.1626 & 0.1787 & 0.1858  
          & 0.1356 & 0.1547 & 0.1568  \\
      &       & EmbSim   
          & 0.1970 & 0.2018 & 0.2149  
          & 0.1654 & 0.1827 & 0.1883  
          & 0.1326 & 0.1529 & 0.1540  \\
      \cmidrule{3-12}
      &       & DTW      
          & \textbf{0.2115}$^\dagger$ & 0.2004 & 0.2093  
          & \textbf{0.1814}$^\ddagger$ & \textbf{0.1870} & \textbf{0.1914}$^\ddagger$  
          & 0.1477$^\ddagger$ & 0.1592 & 0.1582  \\
      &       & Jaccard  
          & 0.2080$^\dagger$ & 0.2129$^\dagger$ & 0.2135  
          & 0.1701 & 0.1860 & 0.1865  
          & 0.1491$^\ddagger$ & 0.1582 & 0.1589  \\
      &       & LCS      
          & \textbf{0.2115}$^\dagger$ & \textbf{0.2149}$^\dagger$ & \textbf{0.2252}$^\dagger$  
          & 0.1651 & 0.1819 & 0.1877  
          & \textbf{0.1596}$^\ddagger$ & 0.1620$^\ddagger$ & \textbf{0.1638}$^\ddagger$  \\
      &       & Time     
          & 0.2004$^\dagger$ & 0.2108 & 0.2198  
          & 0.1705 & 0.1828 & 0.1900$^\ddagger$  
          & 0.1459$^\ddagger$ & \textbf{0.1638}$^\ddagger$ & 0.1617$^\dagger$  \\
    \midrule
    \multirow{12}{*}{\rotatebox[origin=c]{90}{GPT-4o}}
      & \multirow{6}{*}{All}   
        & Random   
          & 0.1403 & 0.1760 & 0.2000  
          & 0.0722 & 0.0956 & 0.1015  
          & 0.0549 & 0.0629 & 0.0674  \\
      &       & EmbSim   
          & 0.1320 & 0.1437 & 0.1735  
          & 0.0705 & 0.0708 & 0.0768  
          & 0.0468 & 0.0430 & 0.0468  \\
      \cmidrule{3-12}
      &       & DTW      
          & 0.2004$^\ddagger$ & 0.2142$^\ddagger$ & 0.2225  
          & 0.1493$^\ddagger$ & 0.1595$^\ddagger$ & 0.1712$^\ddagger$  
          & 0.1054$^\ddagger$ & 0.1072$^\ddagger$ & 0.1145$^\ddagger$  \\
      &       & Jaccard  
          & \textbf{0.2751}$^\ddagger$ & 0.2868$^\ddagger$ & \textbf{0.3089}$^\ddagger$  
          & \textbf{0.1759}$^\ddagger$ & 0.1925$^\ddagger$ & 0.2045$^\ddagger$  
          & 0.1330$^\ddagger$ & 0.1473$^\ddagger$ & 0.1508$^\ddagger$  \\
      &       & LCS      
          & 0.2668$^\ddagger$ & \textbf{0.2909}$^\ddagger$ & 0.2965$^\ddagger$  
          & 0.1712$^\ddagger$ & \textbf{0.1973}$^\ddagger$ & \textbf{0.2088}$^\ddagger$  
          & \textbf{0.1365}$^\ddagger$ & \textbf{0.1554}$^\ddagger$ & \textbf{0.1596}$^\ddagger$  \\
      \cmidrule{2-12}
      & \multirow{7}{*}{User}  
        & Random   
          & 0.3181 & 0.3469 & 0.3522  
          & 0.2406 & 0.2806 & 0.2945  
          & 0.1792 & 0.2014 & 0.2070  \\
      &       & EmbSim   
          & 0.3165 & \textbf{0.3545} & \textbf{0.3587}  
          & 0.2397 & 0.2739 & 0.2947  
          & 0.1648 & 0.1990 & \textbf{0.2112}  \\
      \cmidrule{3-12}
      &       & DTW      
          & 0.3248 & 0.3504 & \textbf{0.3587}  
          & 0.2517$^\ddagger$ & 0.2786 & 0.2969  
          & 0.1861 & 0.2050 & 0.2091  \\
      &       & Jaccard  
          & 0.3276 & 0.3497 & 0.3435  
          & 0.2661$^\ddagger$ & \textbf{0.2935}$^\ddagger$ & 0.3006  
          & 0.1924$^\ddagger$ & 0.2067 & 0.2057  \\
      &       & LCS      
          & \textbf{0.3317}$^\dagger$ & 0.3497 & 0.3580  
          & \textbf{0.2681}$^\ddagger$ & 0.2931$^\ddagger$ & 0.2960  
          & \textbf{0.1941}$^\ddagger$ & 0.2011 & 0.2053  \\
      &       & Time     
          & 0.3179 & 0.3449 & 0.3490  
          & 0.2612$^\ddagger$ & 0.2904$^\ddagger$ & \textbf{0.3047}$^\ddagger$  
          & 0.1826 & \textbf{0.2095}$^\ddagger$ & 0.2067  \\
    \bottomrule
  \end{tabular}
}
\label{tab:main_result}
\end{table}


\section{Results}

\subsection{Comparison of Demonstration Selection Methods}

\noindent
\textbf{Which demonstration selection strategy improves accuracy?}
Table \ref{tab:main_result} compares heuristic demonstration selection methods (DTW, Jaccard, and LCS) with the baselines (Random and EmbSim), showing that heuristic approaches consistently outperform the baselines across all settings in the case of using historical data from all users (All). For instance, on the NYC dataset with Qwen-2.5-7B-Instruct, ACC@1 increases from 0.1320 (EmbSim) to 0.1500 (DTW), 0.1790 (Jaccard), and 0.1776 (LCS). Similar trends are observed in the TKY and CA datasets and experiments using GPT-4o. 

These results confirm that explicitly leveraging spatiotemporal relationships enhances demonstration selection, leading to more effective ICL for next POI prediction. A key limitation of embedding-based selection (EmbSim) is that it does not explicitly incorporate spatiotemporal information when creating embeddings, which can result in suboptimal representations for next POI prediction.


\noindent
\textbf{Does user-based filtering improve accuracy?}
Table~\ref{tab:main_result} shows that the user-based filtering further enhances performance across all settings e.g., on the NYC dataset in the five-shot setting with Qwen-2.5-7B-Instruct, LCS-based selection achieves 0.2115 when combined with user-based filtering, compared to 0.1776 without it.
These findings indicate that incorporating user-specific mobility patterns helps retrieve more relevant demonstrations, leading to better predictions.

\begin{figure}[tb]
\centering
\includegraphics[width=\linewidth]{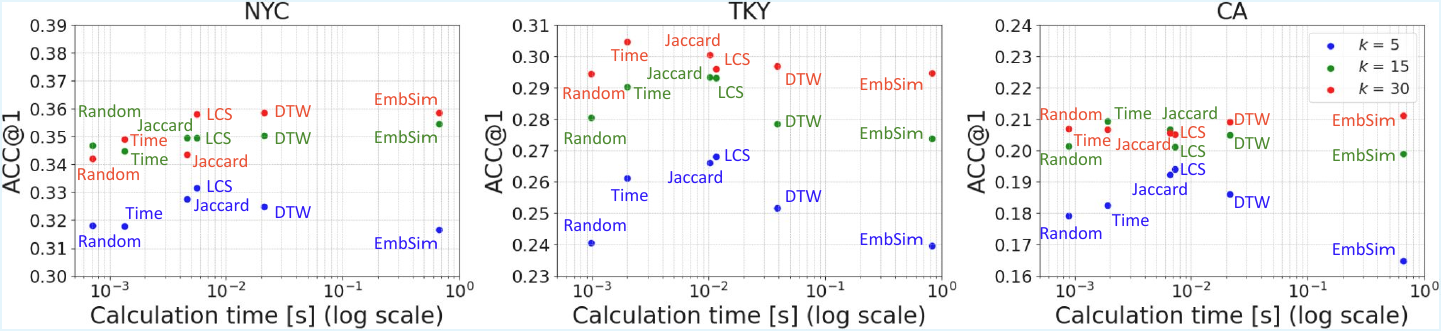}
\vspace{-0.1cm}
\caption{Comparison of demonstration selection computational cost and ACC@1 by each demonstration selection method under GPT-4o with user-filtering settings.}
\label{fig:cost}
\end{figure}

\noindent
\textbf{Which demonstration selection method performs best in terms of both computational cost and accuracy?}
Figure~\ref{fig:cost} compares the computational cost and prediction accuracy of each demonstration selection method.
The computational cost, measured as the average time per test instance, includes both the similarity computation between the test instance and each trajectory in the demonstration pool, as well as the time required for selecting the demonstrations.
For the EmbSim method, the time for encoding the text-representation of POI visit history into an embedding vector is also included; however, once embeddings are computed for an instance, they are reused to avoid redundant computation.
Note that the computation time for demonstration selection does not depend on the number of demonstrations $k$.

As shown in Figure~\ref{fig:cost}, under the $k = 5$ setting, Jaccard and LCS provide the best trade-off between accuracy and computational cost.
In the NYC and CA datasets, when the number of demonstrations $k$ is 15 or more, most of the user’s past trajectories are already included in the demonstration set, resulting in little difference in prediction accuracy between $k = 15$ and $k = 30$.
In contrast, in the TKY dataset, demonstration selection remains important even when $k = 15$.
In this case, Jaccard and LCS again outperform other methods in terms of both accuracy and efficiency.
Although EmbSim incurs high computational costs due to the embedding calculations, it fails to select demonstrations that are effective for next POI prediction.

\begin{table}[tb] 
\caption{ACC@1 scores of selected methods from GPT-4o in the user-filtering setting and fine-tuned models. Fine-tuned models' results are reported values extracted from ~\cite{Li_LLM4POI_SIGIR2024}. McNemar’s test is not applied to this table, as the results of the existing fine-tuned models are not publicly available.}
\centering
\small
\resizebox{\linewidth}{!}{
\begin{tabular}{@{}clccccccccc@{}}
\toprule
 &  & \multicolumn{3}{c}{\textbf{NYC}} & \multicolumn{3}{c}{\textbf{TKY}}  & \multicolumn{3}{c}{\textbf{CA}} \\
\cmidrule(lr){3-5} \cmidrule(lr){6-8} \cmidrule(lr){9-11}
 & & \multicolumn{1}{c}{5} & \multicolumn{1}{c}{15} & \multicolumn{1}{c}{30} & \multicolumn{1}{c}{5} & \multicolumn{1}{c}{15} & \multicolumn{1}{c}{30}  & \multicolumn{1}{c}{5} & \multicolumn{1}{c}{15} & \multicolumn{1}{c}{30} \\
\midrule
\multirow{6}{*}{\rotatebox[origin=c]{90}{GPT-4o}}
    & DTW+User  & 0.3248 & \bf{0.3504} & \bf{0.3587} & 0.2517 & 0.2786 & 0.2969 & 0.1861 & 0.2050 & 0.2091  \\
    & Jaccard+User  & 0.3276 & 0.3497 & 0.3435 & 0.2661 & \bf{0.2935} & 0.3006 & 0.1924 & 0.2067 & 0.2057 \\
    & LCS+User  & \bf{0.3317} & 0.3497 & 0.3580 & \bf{0.2681} & 0.2931 & 0.2960 & \bf{0.1941} & 0.2011 & 0.2053  \\
    & Time+User & 0.3179 & 0.3449 & 0.3490 & 0.2612 & 0.2904 & \bf{0.3047} & 0.1826 & \bf{0.2095} & \bf{0.2067} \\
    \cmidrule{2-11}
    & LLM-Mob  & 0.2979 & 0.3283 & 0.3366 & 0.2212 & 0.2599 & 0.2671 & 0.1753 & 0.1962 & 0.1997 \\
\midrule
\multicolumn{2}{l}{\textbf{Existing Fine-tuned Models}}  \\
\midrule
    & GETNext~\cite{Yang_GETNext_SIGIR2022} & \multicolumn{3}{c}{0.2435} & \multicolumn{3}{c}{0.2254} & \multicolumn{3}{c}{0.1357} \\
    & STHGCN~\cite{Yan_STHGCN_SIGIR2023}  & \multicolumn{3}{c}{0.2734}  & \multicolumn{3}{c}{0.2950} & \multicolumn{3}{c}{0.1730} \\
    & LLM4POI~\cite{Li_LLM4POI_SIGIR2024} & \multicolumn{3}{c}{0.3372} & \multicolumn{3}{c}{0.3035} & \multicolumn{3}{c}{0.2065} \\
\bottomrule
\end{tabular}
}
\label{tab:aginst_supervised_llm-mob}
\end{table}

\subsection{Comparison between fine-tuned models and ICL-based method}

\noindent
\textbf{Does in-context learning outperform fine-tuned models?}
Table~\ref{tab:aginst_supervised_llm-mob} shows best-performing demonstration selection strategies achieve results comparable to or better than fine-tuned models.
On the TKY and CA datasets, our best ICL models match the performance of fine-tuned models, while on the NYC dataset, they outperform fine-tuning-based models. 

For instance, the highest ACC@1 score on the NYC dataset is 0.3587 using GPT-4o with DTW or EmbSim and 30 demonstrations.
In contrast, the best-performing fine-tuned model, LLM4POI, achieves 0.3372, despite requiring extensive training on labeled data.
These results suggest that proper demonstration selection allows ICL to compete with, and even surpass, traditional fine-tuning-based approaches. 

\noindent
\textbf{How do the formats to represent demonstrations affect performance?}
ICL-based methods with heuristic demonstration selection and few-shot prompting outperform LLM-Mob, an existing ICL-based approach~\cite{wang_LLMMob_2024}.
Among these methods, Time+User is the most similar to LLM-Mob, as it selects the latest $k$ days of check-in sequences from a user’s history.
The key difference lies in how demonstrations are structured in the prompt.
LLM-Mob simply lists check-in records as tuples (e.g., ``(01:22 PM, Wednesday, 2236, Train Station)’’), while our approach structures demonstrations, as shown in Figure~\ref{fig:prompt}.
The model with our representation (Time+User in Table~\ref{tab:aginst_supervised_llm-mob}) achieves 0.3179 while LLM-Mob achieves only 0.2979 in the settings with GPT-4o in the five-shot setting on the NYC dataset.
The performance improvement achieved by using the few-shot prompting format is also observed in the TKY and CA datasets. This demonstrates that ICL contributes to performance gains in the next POI prediction task, as with other recommendation tasks.

\subsection{Detailed Analysis}

\begin{figure*}[tb]
\centering
\includegraphics[width=\linewidth]{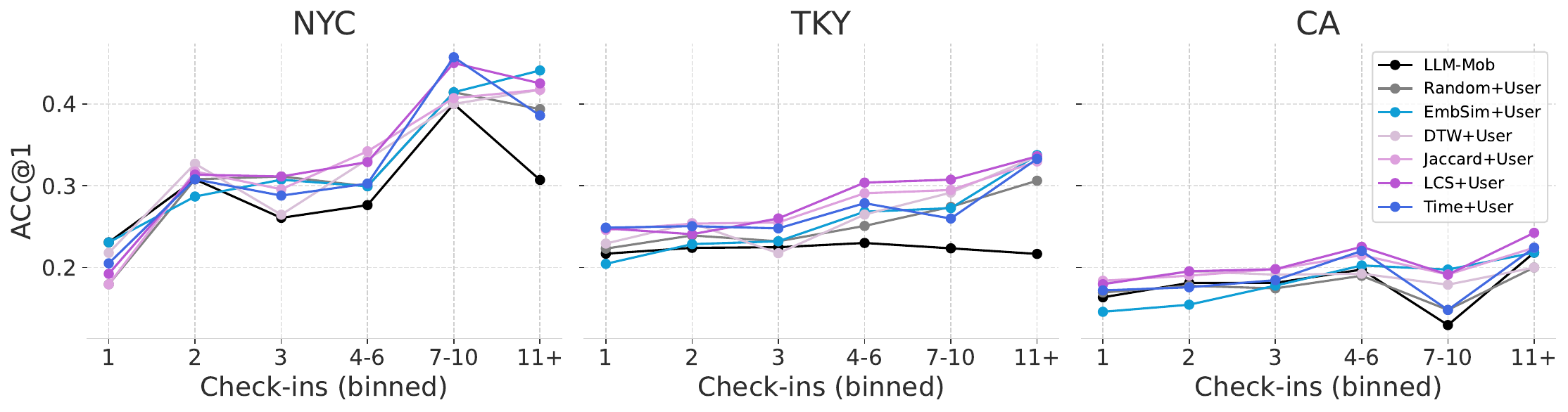}
\vspace{-0.1cm}
\caption{ACC@1 performance of each method across different numbers of current check-ins under the user-filtering setting using GPT-4o.}
\label{fig:checkins_num}
\end{figure*}

\noindent
\textbf{How does the number of current check-ins affect performance?}
Figure~\ref{fig:checkins_num} shows the ACC@1 scores of each demonstration selection method across different numbers of current check-ins.
Overall, we observe that as the number of current check-ins increases, the ACC@1 scores also tend to improve.
This is likely because a longer check-in history makes the user’s movement pattern more explicit, allowing the models to recommend more appropriate POIs.
In contrast, LLM-Mob demonstrates a unique trend: its accuracy decreases as the number of current check-ins increases.
For LLM-Mob, extracting the necessary information from the check-in sequence becomes increasingly difficult as the length grows.
LCS+User and Jaccard+User exhibit relatively high ACC@1 scores across the board.
When there is only a single check-in, the user's movement pattern cannot be captured, making it difficult to select informative demonstrations.
However, as the number of check-ins increases, these selection methods can select demonstrations with useful movement patterns, thereby improving prediction accuracy.

\begin{figure*}[tb]
\centering
\includegraphics[width=\linewidth]{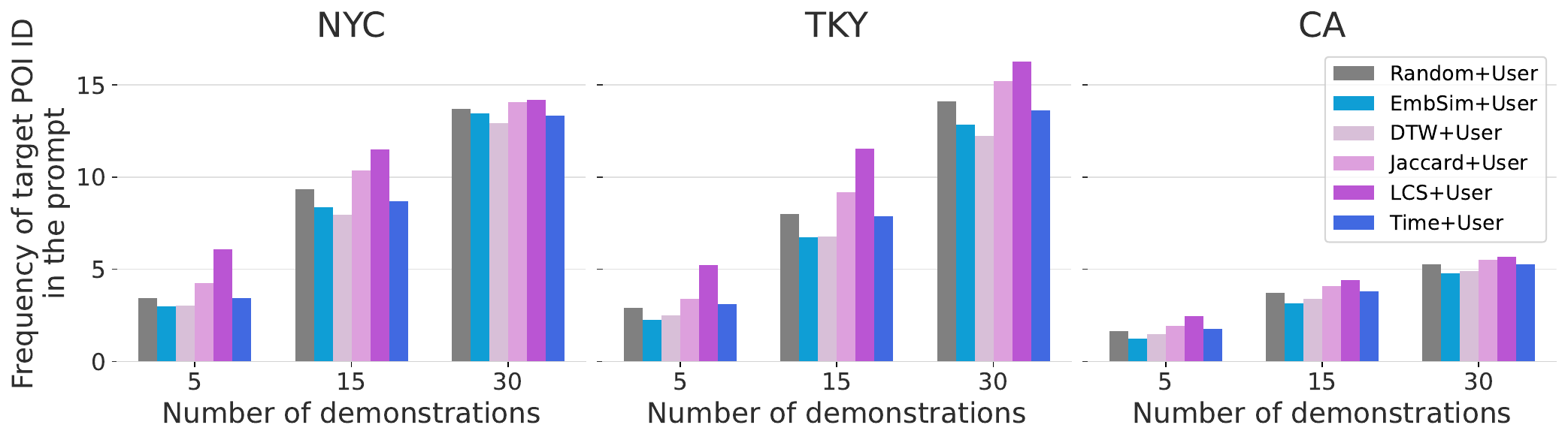}
\vspace{-0.1cm}
\caption{Comparison of the number of target POI labels included as demonstrations in the prompt. Our proposed method efficiently includes the correct labels, which is especially prominent when the number of demonstrations is small.}
\label{fig:poi_num}
\end{figure*}

\noindent
\textbf{How well does our method capture relevant POI information in the prompt?} 
Figure~\ref{fig:poi_num} illustrates the number of target POI ID included in the demonstrations selected by each demonstration selection method. As shown in the figure, the Jaccard and LCS methods include a greater number of target POI ID in the prompt. The Pearson correlation coefficients between the number of target POI IDs appearing in the prompt and ACC@1 with GPT-4o for $k = 5,15,30$ are as follows: 0.7036, -0.0905, and -0.0491 for NYC, 0.7553, 0.7858, and 0.0552 for TKY, and 0.8125, 0.1916, and -0.9569 for CA. 
These results indicate that correctly including the ground-truth POI in the selected demonstrations is crucial when $k$ is small relative to the user's check-in history ($k=5$ in NYC, $k=5,15$ in TKY, $k=5$ in CA).
This also shows that heuristic demonstration selection methods retrieve relevant information more effectively than random selection and embedding-based selection.

\section{Conclusions}
This paper comprehensively evaluates demonstration selection methods for next-POI prediction. Experiments show simpler heuristics--DTW, Jaccard, and LCS--consistently outperform random and embedding-based selections.
This advantage is particularly evident in settings with fewer demonstrations, where choosing the most informative examples is critical due to prompt length limits. 
Embedding-based methods poorly capture movement patterns and incur high costs to compute embeddings.
In contrast, methods based on Jaccard and LCS offer a practical alternative by selecting informative demonstrations with much lower computation, while achieving better accuracy.
These findings suit real-world applications where efficiency and effectiveness are crucial.
As future work, we aim to develop effective demonstration selection methods using other users' historical data to address the cold-start problem.
Furthermore, 
another important direction is to enhance the performance of models that can run locally for practical deployment.
While this study did not utilize POI names, we believe that incorporating them could be beneficial in LLM-based prediction tasks. Although traditional next POI prediction approaches have largely overlooked this feature, future work includes building datasets that include POI names and evaluating whether such semantic information can further improve performance.

\section*{Acknowledgment}
This work was supported by Japan Society for the Promotion of Science under KAKENHI Grant Number 24K20850, and 
was based on results obtained from a project, Programs for Bridging the gap between R\&D and the IDeal society (society 5.0) and Generating Economic and social value (BRIDGE)/Practical Global Research in the AI × Robotics Services, implemented by the Cabinet Office, Government of Japan.

\bibliographystyle{splncs04}
%




\bibliography{paper}

\end{document}